\documentclass[aps,prx,onecolumn,superscriptaddress,nofootinbib,longbibliography,notitlepage]{revtex4-2}

\usepackage[utf8]{inputenc}
\usepackage[T1]{fontenc}
\usepackage{lmodern}
\usepackage{amsmath,amssymb,amsfonts}
\usepackage{bm}
\usepackage{graphicx}
\usepackage{microtype}
\usepackage[hidelinks]{hyperref}
\usepackage[capitalise]{cleveref}
\usepackage{xcolor}
\usepackage{siunitx}
\usepackage{booktabs}
\usepackage{enumitem}
\usepackage{tikz}
\usetikzlibrary{positioning,arrows.meta,fit}
\usepackage{tabularx}

\begin{document}

\title{Deep Learning Theory through the Lens of Statistical Physics:\\ A Topical Review}

\author{Wei Huang}
\email{wei.huang.vr@riken.jp}
\affiliation{RIKEN Center for Advanced Intelligence Project (AIP), Tokyo, Japan}
\affiliation{The Institute of Statistical Mathematics, Tokyo, Japan}

\author{Youjin Deng}
\affiliation{Hefei National Research Center for Physical Sciences at the Microscale and School of Physical Sciences, University of Science and Technology of China, Hefei 230026, China}
\affiliation{Hefei National Laboratory, University of Science and Technology of China, Hefei 230088, China}
\affiliation{Shanghai Research Center for Quantum Science and CAS Center for Excellence in Quantum Information and Quantum Physics, University of Science and Technology of China, Shanghai 201315, China}

\begin{abstract}
The recent success of foundation models and deep learning has far outpaced our theoretical understanding of how large-scale neural networks learn and generalize. Statistical physics provides a natural language for describing such high-dimensional, nonlinear, and strongly coupled systems, offering concepts such as mean-field theory, phase transitions, energy landscapes, and stochastic dynamics. In this topical review, we adopt a statistical-physics way of thinking---from concepts including phases and criticality, and free-energy viewpoints, to methods such as kernel and Gaussian-process limits, mean-field and dynamical mean-field theories, random-matrix tools, and Fokker--Planck formalisms, and to a mindset emphasizing universality, scaling, and tractable limits with explicit validity regimes. We survey recent progress on understanding modern deep learning through this lens, including mean-field analyses of wide networks, spectral and geometric perspectives on loss landscapes, stochastic-gradient-based training viewed as a dynamical process, and phase-transition analogies that organize phenomena such as overparameterization, generalization, and representation change. Our goal is to synthesize these threads into a coherent framework that clarifies what each approximation explains, when it is expected to hold, and how apparently competing narratives relate or diverge. Together, these perspectives form a unified physical framework that views representation learning as a collective phenomenon in high-dimensional systems, bridging deterministic and stochastic regimes of gradient-based optimization.
\end{abstract}

\keywords{deep learning theory; statistical physics; mean-field theory; neural tangent kernel; loss landscape; stochastic gradient descent; phase transitions; non-equilibrium dynamics}

\maketitle

\begingroup
\renewcommand\thefootnote{}
\footnotetext{This is the authors' version of the article published in Front.\ Phys.\ \textbf{22}(1), 011302 (2027), \url{https://doi.org/10.15302/frontphys.2027.011302}. It contains minor corrections relative to the published version.}
\endgroup

\section{Introduction}
\label{sec:intro}

The past decade has witnessed the remarkable rise of deep learning and, more recently, foundation models that attain near-human performance across language, vision, and multimodal domains \cite{lecun2015deep,goodfellow2016deep,brown2020language,radford2021learning}. Yet despite these empirical successes, our theoretical understanding of how large-scale neural networks learn, form representations, and generalize remains incomplete. Classical statistical learning theory provides valuable asymptotic insights and worst-case guarantees, but it often fails to explain the practical behavior of highly overparameterized, strongly nonlinear models trained by modern optimization pipelines \cite{zhang2016understanding,belkin2019reconciling}. This mismatch has motivated an active search for analytical frameworks that can capture the collective behavior of neural networks in high dimensions.

Statistical physics offers a natural language for this purpose: it was developed to describe macroscopic regularities emerging from many interacting degrees of freedom, and it provides concepts and tools that transfer naturally to learning systems. In the context of deep learning, these include control parameters, phase-transition and criticality analogies, energy- (or free-energy-) landscape viewpoints, and stochastic dynamics. Equally important is the accompanying mindset---universality, scaling, and solvable limits with explicit validity regimes---which encourages identifying which phenomena are robust across architectures and which depend sensitively on modeling or optimization details. This review adopts that statistical-physics way of thinking as an organizing lens for modern deep learning theory.

Early connections between neural computation and statistical mechanics were established via spin-glass-inspired models of associative memory \cite{hopfield1982neural} and analyses of random recurrent networks with chaotic dynamics \cite{sompolinsky1988chaos}. Subsequent developments extended these ideas to perceptrons and multilayer networks using replica and mean-field methods to study capacity, generalization, and landscape structure \cite{seung1992statistical,engel2001statistical}. These classical statistical-mechanics approaches to neural networks are synthesized in several recent textbooks~\cite{huang2021statistical,helias2020statistical}. In recent years, these physical intuitions have been revived and expanded for modern deep networks, linking high-dimensional inference, optimization, and representation learning---often in regimes that are inherently out of equilibrium \cite{advani2020high,mei2018mean,bordelon2022self}.

A first pillar of this revival is the static viewpoint provided by loss-landscape geometry and spectral structure. The training objective defines a high-dimensional, nonconvex surface whose qualitative features---ridges, valleys, saddles, and degeneracies---invite comparisons with disordered systems and motivate phase-transition analogies \cite{choromanska2015loss,advani2020high}. Random matrix theory complements this view by characterizing spectra of Jacobians and Hessians that govern conditioning, curvature, and stability of learning dynamics \cite{sagun2017empirical,pennington2018emergence,martin2021implicit}. These landscape- and spectrum-based perspectives provide a compact set of empirical and theoretical targets for any mechanistic theory of learning in deep networks.

A second pillar is the development of controlled, solvable limits in which neural networks admit tractable statistical descriptions. In the infinite-width limit under appropriate scalings, random networks converge to Gaussian processes, yielding the Neural Network Gaussian Process (NNGP) view of randomly initialized models \cite{neal2012bayesian,lee2017deep,matthews2018gaussian}. Linearizing training dynamics around initialization leads to the Neural Tangent Kernel (NTK) regime, where gradient flow behaves like kernel regression with a fixed kernel \cite{jacot2018neural}. These theories explain concentration phenomena and provide sharp predictions for optimization and generalization in wide networks, but they also characterize a ``lazy'' regime in which features change little during training. To move beyond this regime and model representation change explicitly, mean-field and dynamical formulations track the evolution of neuron distributions and their interactions, typically through measure-valued dynamics and self-consistent closures \cite{mei2018mean,chizat2018global,rotskoff2018parameters}.

\begin{figure*}[t]
\centering
\includegraphics[width=0.95\linewidth]{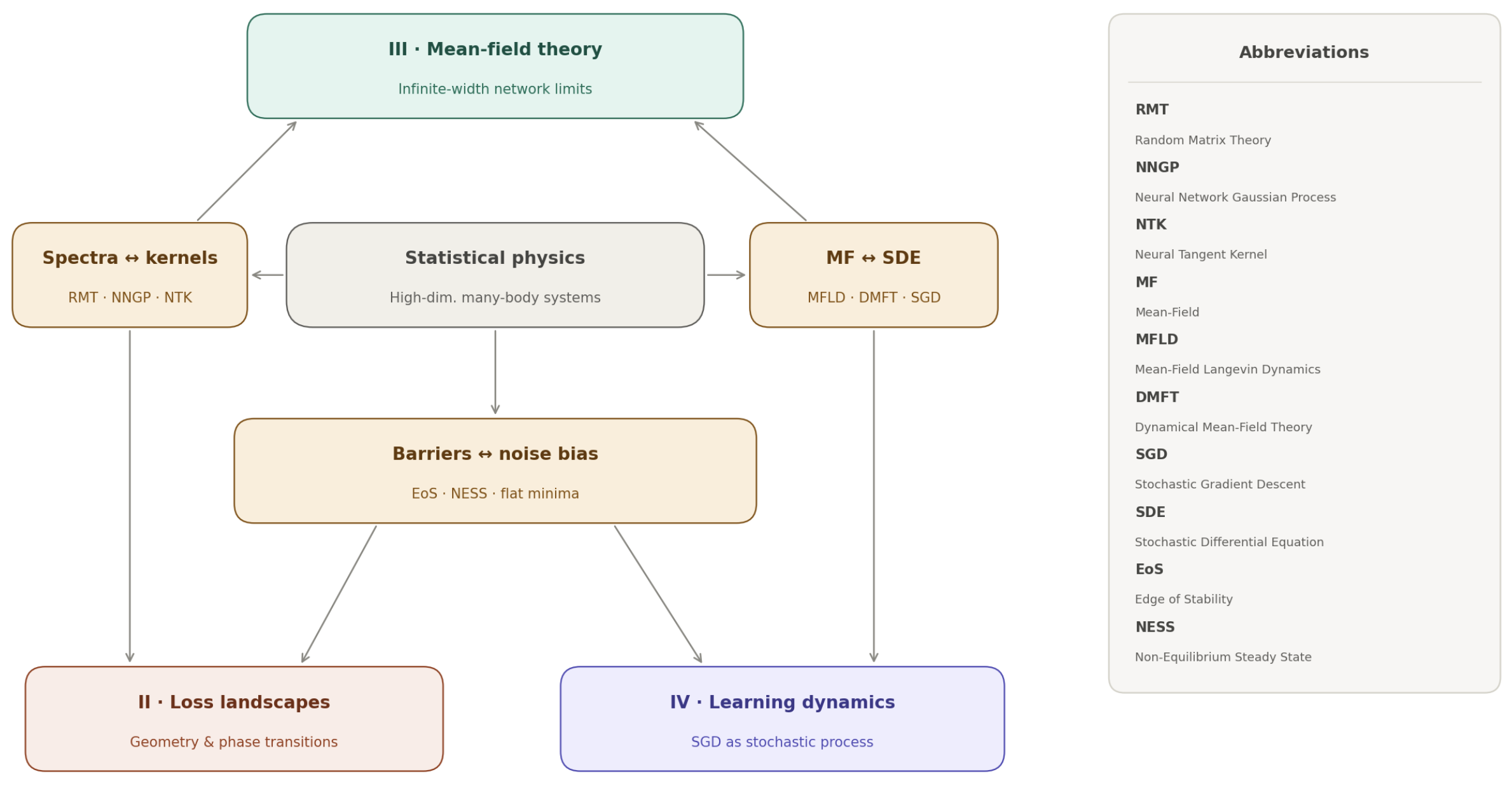}
\caption{Relationship map of the review. Three theoretical pillars---loss-landscape geometry (Sec.~\ref{sec:loss_landscape}), infinite-width mean-field limits (Sec.~\ref{sec:mean_field}), and stochastic learning dynamics (Sec.~\ref{sec:sde})---are connected by shared statistical-physics concepts: spectral universality, mean-field stochastic equations, and noise-driven escape from metastable minima. Abbreviations are listed in the sidebar; see also Table~\ref{tab:dictionary} for the full statistical-physics dictionary.}
\label{fig:roadmap}
\end{figure*}

A third pillar focuses on training-time dynamics and the role of stochasticity. From a statistical-physics standpoint, stochastic gradient-based training can be modeled as a stochastic dynamical system that explores a complex landscape under state-dependent noise \cite{mandt2017stochastic,mignacco2021stochasticity}. This viewpoint suggests temperature-like effective parameters that modulate exploration--exploitation trade-offs and helps explain why optimization and generalization are inseparable in practice. Importantly, because the noise is generally anisotropic and depends on parameters and data, the resulting dynamics need not admit an equilibrium Gibbs description, motivating non-equilibrium viewpoints for modern learning systems. Together, these three pillars and their interconnections are summarized in Figure~\ref{fig:roadmap}.

Beyond classical architectures, recent advances in generative modeling and large-scale training introduce additional statistical-physics analogies. Diffusion and score-based generative models can be interpreted as learning reverse-time dynamics of a forward noising process that transports data to noise, connecting generation to Langevin-type sampling and reverse-time Fokker--Planck descriptions under suitable discretizations and assumptions \cite{sohl2015deep,ho2020denoising,song2019generative,song2020score}. In parallel, empirical scaling laws in foundation models relate performance predictably to data, model size, and compute across wide ranges \cite{kaplan2020scaling,hoffmann2022training}. While the underlying mechanism is still under active investigation, the apparent self-similarity and regularity invite comparisons to universality and near-critical behavior in statistical systems. Together, these developments broaden the interface between deep learning and statistical physics and motivate a unified view of modern models as high-dimensional dynamical systems.

\subsection{Related reviews and scope}
\label{sec:scope}

Recent perspectives have notably revisited the foundational connections between statistical mechanics and neural computation, particularly inspired by the historical interplay between spin glasses and associative memory. For example, Pei~\cite{pei2026statistical} provides a broad overview of these classical analogies, exploring the topological similarities between Ising-type models (such as Hopfield networks and Boltzmann machines) and biological neurons, with a specific outlook on their implications for neuromorphic and quantum computing. In contrast, the present review focuses on the theoretical frameworks describing modern deep learning architectures trained via gradient-based optimization. Rather than revisiting the static energy landscapes of classical associative memories, we concentrate on the collective dynamics of high-dimensional, multi-layer networks in the overparameterized regime. We survey rigorous analytical limits---ranging from the static NNGP and NTK descriptions to Dynamical Mean-Field Theories (DMFT) that capture active feature learning---and analyze the training process itself as a non-equilibrium stochastic dynamical system. Our goal is to provide a unified physical picture of how deep networks generalize and evolve representations, bridging the gap between idealized infinite-width limits and the finite-width, stochastic reality of practical training.

Our work builds upon the foundational survey by Bahri~et~al.~\cite{bahri2020statistical}, which comprehensively bridged condensed matter physics and deep learning by analyzing static error landscapes, spin-glass analogies, and the order-to-chaos transition in signal propagation at initialization. Since their publication, however, the theoretical frontier has shifted from characterizing these static properties and the ``lazy'' training regime toward understanding the rich dynamics of feature learning where internal representations actively evolve. We complement their perspective by focusing on self-consistent dynamical frameworks---such as Dynamical Mean-Field Theory (DMFT)---that capture this evolution beyond the frozen kernel limit. Furthermore, while previous works touched upon equilibrium statistical mechanics, we place a stronger emphasis on the non-equilibrium stochastic thermodynamics of training and on the anisotropic, state-dependent noise structure of SGD.

We also distinguish our contribution from the comprehensive review by Carleo~et~al.~\cite{carleo2019machine}, which surveyed the broad interface between machine learning and the physical sciences. While their work provided a seminal overview of how machine learning techniques accelerate discovery in fields ranging from particle physics to quantum many-body systems (``AI for Science''), our review focuses exclusively on the complementary direction: applying statistical physics concepts to build a rigorous theoretical understanding of deep learning itself (``Physics of AI''). Furthermore, since the publication of Carleo~et~al., the field has matured from analyzing static architectures and equilibrium properties to characterizing the non-equilibrium kinetics of training. We specifically address these post-2019 developments, reviewing how dynamical mean-field theories describe the crossover between the ``lazy'' and ``rich'' (feature-learning) regimes. Although a systematic treatment of generative models is beyond our scope, recent statistical-physics analyses of diffusion models have clarified their dynamical regimes and non-equilibrium structure~\cite{biroli2024dynamical,yu2025nonequilbrium}.

\subsection{Organization}
\label{sec:org}

In this topical review, we synthesize these complementary perspectives into a coherent statistical-physics framework for deep learning, emphasizing both what each approach explains and the regimes where its assumptions are expected to hold. We highlight how tools such as mean-field and dynamical mean-field theories, random matrix methods, landscape analyses, and stochastic-process formalisms can be used to interpret learning, optimization, and generalization in modern neural networks, and we stress finite-width and finite-data effects as the key frontier beyond idealized limits. The review is organized as follows. Section~\ref{sec:loss_landscape} begins with landscape geometry, spectral structure, and phase-transition analogies. Section~\ref{sec:mean_field} then introduces solvable wide-network limits, including NNGP and NTK descriptions and mean-field formulations that address feature learning beyond lazy regimes. Section~\ref{sec:sde} focuses on learning dynamics as stochastic processes, discussing SDE and non-equilibrium viewpoints of SGD, the structure of gradient noise, and metastable first-passage dynamics. We conclude by outlining open challenges, including bridging infinite- versus finite-width regimes and reconciling equilibrium-inspired pictures with the realities of non-equilibrium training dynamics.

\section{Loss landscapes and phase-transition analogies}
\label{sec:loss_landscape}

Deep neural network training can be viewed as the search for low-energy configurations on a high-dimensional, non-convex loss surface. This perspective invites a statistical-physics description in which the empirical risk plays the role of an energy function, the parameters are microscopic degrees of freedom, and optimization algorithms define dynamical rules that explore the resulting energy landscape. While the loss landscape is not a physical Hamiltonian in a strict sense (e.g., it is data-dependent and typically non-stationary under changing training protocols), the analogy has proven fruitful: it motivates the use of concepts such as critical points and saddles, metastability and barriers, symmetry and symmetry breaking, and even phase-transition-like phenomena that emerge as model size, data size, regularization strength, or training time are varied. To systematically organize these parallels, we summarize the key correspondences between deep learning and statistical physics in Table~\ref{tab:dictionary}.

\begin{table*}[t]
\caption{\textbf{A statistical-physics dictionary for deep learning.} This table summarizes the correspondence between key concepts in deep learning and their analogies in statistical mechanics. These mappings provide the conceptual basis for understanding phenomena such as flat minima selection, phase transitions, and non-equilibrium dynamics in overparameterized networks.}
\label{tab:dictionary}
\begin{ruledtabular}
\begin{tabular}{lll}
\textbf{Deep Learning Concept} & \textbf{Statistical Physics Analogy} & \textbf{Physical Interpretation \& Role} \\
\hline
Loss function $\mathcal{L}({\theta})$ & Energy $H({\sigma})$ &
The potential landscape governing the system's dynamics. \\
& & Global minima correspond to ground states. \\[4pt]

Parameters ${\theta}$ & Microscopic degrees of freedom &
The interacting variables (e.g., spins in a spin glass) \\
& & that define the system's microscopic configuration. \\[4pt]

SGD noise & Effective temperature $T_\mathrm{eff}$ &
Stochastic fluctuations regulating exploration--exploitation. \\[4pt]

Flat minima & High-entropy basin ($-TS$ term) &
Solutions occupying large volume in parameter space, \\
& & entropically favored under SGD's implicit bias toward \\
& & low gradient variance. \\[4pt]

Sharp minima & Low-entropy metastable state &
High-curvature solutions with elevated escape rates; \\
& & SGD noise preferentially destabilizes these due to \\
& & noise--curvature alignment. \\[4pt]

Over-parameterization & Thermodynamic limit &
Fluctuations vanish; macroscopic self-consistent \\
& & descriptions (NNGP, NTK, DMFT) emerge. \\[4pt]

Interpolation threshold & Phase transition point &
Critical point separating underfitting and overfitting \\
& & regimes, marked by diverging estimator variance. \\[4pt]

Lazy-to-rich transition & Symmetry-breaking-like crossover &
Neurons specialize from a disordered (kernel) phase \\
& & to a structured (feature-learning) phase; whether a \\
& & sharp transition exists is model dependent. \\
\end{tabular}
\end{ruledtabular}
\end{table*}

\subsection{Geometric structure of high-dimensional loss landscapes}
\label{sec:geometry}

A central question concerns the nature and distribution of critical points of the empirical risk
\begin{equation}
\label{eq:empirical-risk}
\mathcal{L}(\theta)=\frac{1}{n}\sum_{i=1}^n \ell\!\left(f_\theta(x_i),y_i\right),
\end{equation}
where $\ell$ is the loss function, $\theta$ denotes the parameters, and $\{(x_i,y_i)\}_{i=1}^n$ are training samples. Understanding the geometry of $\mathcal{L}(\theta)$ is central to both optimization and generalization.

Early work drew explicit analogies between deep-network loss surfaces and spin-glass energy landscapes, suggesting that in sufficiently high dimensions, saddle points dominate the landscape while poor local minima become exponentially rare \cite{bray2007statistics,choromanska2015loss}. These results rely on spin-glass analogies whose underlying assumptions---variable independence and parametric redundancy---remain unverified for practical architectures. In this view, optimization difficulty is governed less by trapping in bad minima and more by the abundance of flat or weakly unstable directions. Complementary to the spin-glass intuition, a line of work establishes the absence of suboptimal local minima in analytically tractable settings (notably deep linear networks), suggesting that optimization difficulty may instead arise from saddle structure---potentially including degenerate saddle points in sufficiently deep models~\cite{kawaguchi2016deep}. However, two important caveats apply: the result is restricted to squared loss and does not extend to cross-entropy; and in networks deeper than three layers, degenerate saddle points with no strictly negative Hessian eigenvalue can exist, which can make gradient-based escape more difficult \cite{kawaguchi2016deep,lu2017depth,laurent2018deep}.

\begin{figure}[tbp]
    \centering
    \includegraphics[width=\linewidth]{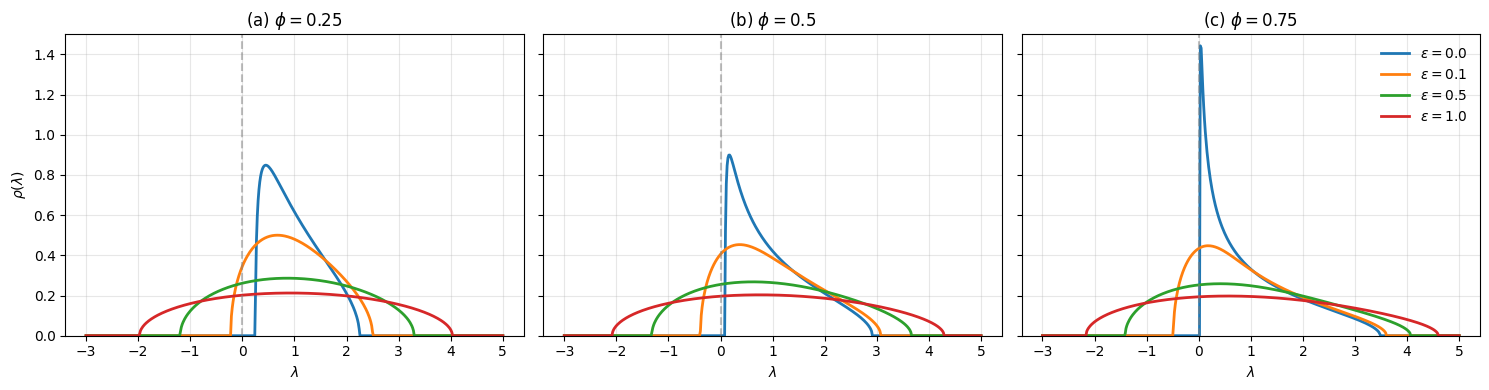}
    \caption{\textbf{Spectral evolution of the Hessian matrix at critical points.} The plots show the spectral density $\rho(\lambda)$ for different parameter-to-data ratios $\phi$ and energy levels $\epsilon$, derived using random matrix theory (reproduced from~\cite{pennington2017geometry}). As the energy $\epsilon$ increases, the spectrum transitions from a bulk with positive outliers (resembling the Marchenko--Pastur distribution, associated with stable minima) to a broader, semi-circular distribution spanning negative eigenvalues (associated with unstable saddle points). This illustrates the landscape's transition from an ordered phase to a disordered phase in the high-dimensional parameter space.}
    \label{fig:hessian_spectrum}
\end{figure}

From a modern perspective, the local geometry around a solution $\theta^\star$ is characterized by the Hessian $H(\theta^\star)=\nabla^2_\theta \mathcal{L}(\theta^\star)$. Empirical studies of trained deep networks consistently reveal a spectrum composed of a small number of large outlier eigenvalues and a broad bulk concentrated near zero \cite{sagun2017empirical,pennington2018emergence,mahoney2019traditional,ghorbani2019investigation}. Crucially, the shape of this spectrum evolves with the energy level of the solution, as illustrated in Figure~\ref{fig:hessian_spectrum}: high-error states are characterized by a proliferation of negative eigenvalues (high index), resembling random Gaussian landscapes, while low-error solutions typically exhibit a bulk distribution bounded away from zero (except for degenerate modes) with positive outliers, often compared with the Marchenko--Pastur (MP) law. While the spiked covariance model (MP bulk plus finite-rank outliers) provides a useful qualitative framework, the MP law applies rigorously only to i.i.d.\ sample covariance matrices; in practice, trained-network Hessians exhibit heavier-tailed bulks than MP predicts \cite{papyan2018full}. Empirical and rigorous analyses further indicate that SGD training trajectories align preferentially with the outlier eigenspaces, directly linking spectral structure to learning dynamics \cite{gur2018gradient,arous2023high}. Random matrix theory provides a natural framework for describing this structure, with the bulk exhibiting universal features and the outliers encoding structured signal components \cite{pennington2017geometry,couillet2011random,baskerville2022appearance}. This spectral decomposition is widely used to analyze training stability and sensitivity to perturbations, and to quantify the effective dimensionality of learned solutions (e.g., via sharpness, conditioning, and related spectral measures) \cite{pennington2018emergence,mahoney2019traditional,couillet2011random}.

\begin{figure}[tbp]
    \centering
    \includegraphics[width=0.5\linewidth]{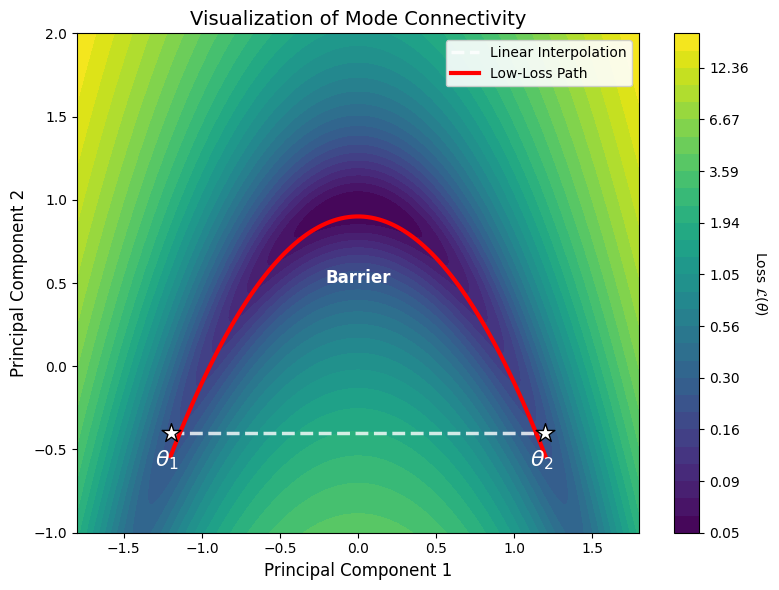}
    \caption{\textbf{Schematic visualization of mode connectivity in the loss landscape.} The plot illustrates a 2D projection of the loss surface containing two independent solutions, $\theta_1$ and $\theta_2$. A linear interpolation between them (white dashed line) traverses a region of high loss (energy barrier). However, consistent with the phenomenon of mode connectivity~\cite{garipov2018loss,draxler2018essentially}, a non-linear path exists (solid red curve) along which the loss remains consistently low. This topology suggests that distinct minima belong to a single connected manifold of solutions, facilitating exploration by stochastic optimization algorithms.}
    \label{fig:mode_connectivity}
\end{figure}

In highly overparameterized regimes, global minimizers are typically not isolated points but form connected manifolds due to parameter redundancies and functional equivalences~\cite{baldassi2016unreasonable,freeman2016topology}. Crucially, empirical and theoretical works have established the phenomenon of \textit{mode connectivity}, revealing that distinct minima found by stochastic optimization are connected by continuous paths of near-zero loss, as visualized in Figure~\ref{fig:mode_connectivity}~\cite{garipov2018loss,draxler2018essentially,zhou2023going}. On the theoretical side, Nguyen~\cite{nguyen2019connected} gave a rigorous proof that for sufficiently overparameterized deep networks with piecewise-linear activations, every sublevel set of the empirical loss is connected and unbounded, implying that all global minima lie within a single connected manifold rather than in isolated valleys. From a physical perspective, these connected valleys arise from the symmetries of the network---continuous rescaling invariances and discrete permutation invariances among hidden units---which generate extensive neutral subspaces in the landscape. This structure is reflected in the Hessian spectrum as a dense cluster of zero eigenvalues. Consequently, the loss landscape is more accurately described as a collection of percolating wide valleys rather than discrete basins, with optimization trajectories confined to these low-curvature subspaces.

\subsection{Flatness, sharpness, and the edge of stability}
\label{sec:flatness}

A widely discussed heuristic links generalization performance to the ``flatness'' of the basin surrounding a trained solution. Intuitively, flatter minima correspond to models that are robust to parameter perturbations, suggesting that the learned function is less sensitive to the precision of the weights and effectively lower in complexity~\cite{hochreiter1997flat,keskar2016large}. From this perspective, the local geometry---characterized, for instance, by the spectrum of the Hessian---is taken as a proxy for the generalization gap.

However, the relationship between geometrical flatness and generalization is subtle. Naive measures of sharpness are sensitive to parameterization; for example, in networks with positively homogeneous activations (e.g., ReLU), weights can be rescaled to arbitrarily alter curvature without changing the function output~\cite{dinh2017sharp}. A statistical-physics interpretation resolves this ambiguity by introducing the concept of \emph{local entropy} or an \emph{effective free-energy landscape}~\cite{baldassi2016unreasonable,chaudhari2019entropy}.
Rather than minimizing the bare loss energy $\mathcal{L}(\theta)$, stochastic optimization is viewed as biasing the measure toward regions of high {local entropy}. 
Concretely, the robust-ensemble construction of Baldassi
et al.~\cite{baldassi2016unreasonable} introduces a
large-deviation construction for the local entropy: configurations are favored when they are surrounded by an exponentially large number of nearby low-loss solutions, i.e., dense ``robust ensembles'', rather than merely because they minimize the bare loss.

Theoretical analyses of discrete neural networks and perceptrons reveal that the loss landscape may be rugged, containing many isolated solutions or narrow minima, a phenomenon already identified in the statistical physics of binary synapses \cite{huang2014origin}. In contrast, the solutions most accessible to stochastic algorithms can form dense clusters, often referred to as robust ensembles \cite{baldassi2021unveiling}.
SGD, driven by anisotropic noise, can be interpreted as being biased toward regions of high local entropy, naturally drifting towards these wide, high-volume basins favored by the large-deviation (local-entropy) measure, even if they are not the global minima of the bare loss. In the context of Bayesian learning, this entropic perspective finds a rigorous counterpart through PAC-Bayes generalization bounds~\cite{mcallester1998some,dziugaite2017computing}: a flat minimum allows the posterior distribution over parameters to retain large variance (and hence large volume in parameter space) while maintaining low expected loss, which directly tightens the bound. This theoretical insight has recently translated into practical algorithms such as Sharpness-Aware Minimization (SAM) methods~\cite{foret2020sharpness,kwon2021asam}, which explicitly minimize a perturbed loss that can be viewed as encouraging robustness to local parameter perturbations.

A closer look at the training dynamics reveals a more nuanced picture that ultimately reinforces the flat-minima narrative. Under full-batch gradient descent (GD), sharpness does not decrease monotonically but instead increases---a phenomenon known as progressive sharpening~\cite{wang2022analyzing}---until the largest Hessian eigenvalue saturates the stability threshold $2/\eta$, a regime called the Edge of Stability (EoS)~\cite{cohen2021gradient}. Subsequent theoretical work has progressively explained this behavior: self-stabilization near EoS arises from cubic terms in the Taylor expansion of the loss~\cite{damian2022self}; gradient flow along the minimum-loss manifold produces the characteristic sharpening trajectory~\cite{arora2022understanding}. More recently, a minimalist analytically tractable model has been constructed that simultaneously exhibits progressive sharpening and the edge-of-stability transition, providing a clean setting in which both phenomena can be studied rigorously~\cite{liu2025minimalist}. Because full-batch GD drives sharpness up to the learning-rate-dependent threshold $2/\eta$, whereas mini-batch noise typically stabilizes training at lower sharpness, SGD-trained solutions tend to be flatter than their GD counterparts at the same learning rate---supporting, rather than contradicting, the entropic view.

The structure of the gradient noise further strengthens this conclusion. The noise covariance of SGD aligns with the Hessian~\cite{haochen2021shape,wu2022alignment}, producing stronger fluctuations along sharper directions. This alignment makes anisotropic SGD noise more effective at selecting flat solutions than isotropic noise would be, as the escape rate from sharp minima is preferentially amplified. SGD can thus be viewed as implicitly reducing the variance of its own stochastic gradients, which---given the noise--curvature alignment---typically favors flat, low-sharpness regions of the landscape, although the direction of this bias can reverse under anisotropic label noise~\cite{xu2026does}.

\subsection{Symmetry, degeneracy, and symmetry breaking}
\label{sec:symmetry}

Deep neural networks possess multiple sources of symmetry that drastically shape the geometry of their loss landscapes. Discrete permutation symmetries of hidden units leave the computed function invariant, while continuous rescaling symmetries arise in networks with positively homogeneous activations (e.g., ReLU). These symmetries generate extensive families of equivalent parameter configurations, leading to a proliferation of degenerate global minimizers and nearly flat directions in parameter space. Crucially, this degeneracy follows a rigorous \textit{embedding principle}: the loss landscape of an overparameterized network topologically contains the critical points of all narrower networks~\cite{zhang2021embedding}. Consequently, local minima in lower-dimensional subspaces effectively act as saddle points or degenerate valleys in the higher-dimensional landscape, facilitating optimization by opening up escape directions.

From a statistical-physics perspective, these degeneracies generate continuous and discrete invariance groups acting on parameter space, analogous to the global symmetries of disordered systems rather than to local gauge symmetries in the strict physical sense. Early landscape analysis revealed the phenomenon of mode connectivity, where distinct minima found by SGD are connected by low-loss paths~\cite{draxler2018essentially,garipov2018loss}. Recent theoretical advances have deepened this understanding by explicitly linking connectivity to symmetry~\cite{entezari2021role,ainsworth2022git,ziyin2023symmetry}. Entezari~et~al.~\cite{entezari2021role} conjectured that accounting for permutation symmetries of hidden units is sufficient to linearly connect any two solutions, and Ainsworth~et~al.~\cite{ainsworth2022git} largely confirmed this conjecture, showing that after re-aligning neurons via permutation, distinct minima found by independent SGD runs become linearly mode-connected with near-zero loss barrier. This suggests that the complex loss landscape may consist of essentially a single connected basin modulo discrete symmetry groups---a profound structural simplification.

By analogy with physical systems, the transition from lazy training to feature learning can be interpreted, in some models, as the emergence of macroscopic order. At exchangeable initialization, neurons are statistically equivalent. In the lazy regime, their parameter displacements remain small and the dynamics are well described by an approximately fixed kernel~\cite{chizat2019lazy}. In feature-learning regimes, individual neurons may instead develop heterogeneous and specialized responses, while the effective tangent kernel evolves during training~\cite{bordelon2022self}. This sample-level organization is reminiscent of symmetry breaking, but it does not remove the exact permutation invariance of the network, and the population law may remain exchangeable. Accordingly, the relevant order parameter and the existence of a sharp phase transition are model dependent. DMFT provides a framework for characterizing such crossovers through evolving correlation and response functions, rather than establishing a universal breaking of neuron-permutation symmetry.

\subsection{Double descent and interpolation thresholds}
\label{sec:double_descent}

One of the most striking empirical phenomena in modern machine learning is the so-called double-descent behavior of test error as a function of model capacity or training time. As capacity increases, generalization error initially decreases following the classical bias--variance tradeoff, then peaks near the interpolation threshold where the training error first reaches zero, and finally decreases again in the highly overparameterized regime~\cite{belkin2019reconciling,advani2020high,nakkiran2021deep,hastie2022surprises}. Its essential structure was already identified in the statistical-physics analysis of perceptron learning as a phase-transition problem~\cite{hertz1989phase}. This nonmonotonic behavior challenges traditional statistical learning intuition and has motivated renewed interest in high-dimensional theories of generalization.

From the viewpoint of loss landscape geometry, the interpolation threshold corresponds to a qualitative change in the structure of the zero-loss level set. Below the threshold, no parameter configuration achieves zero training error, while at the threshold the solution set first appears as a marginal, low-dimensional manifold. Above the threshold, the set of global minima typically expands into a high-dimensional connected region~\cite{cooper2018loss,venturi2019spurious}. This transition in the dimensionality and topology of the zero-loss manifold provides a direct geometric link between double descent and loss landscape structure.

From a statistical-physics perspective, the interpolation threshold can be interpreted as a transition point separating qualitatively distinct regimes of solution geometry. Below the threshold, the learning problem is underparameterized and constraints cannot be simultaneously satisfied. At the threshold, the system becomes marginally satisfiable, leading to enhanced sensitivity to noise and data perturbations, while above the threshold, the solution set typically expands into a high-dimensional manifold of interpolating solutions. This qualitative change bears a close analogy to phase transitions in constraint-satisfaction problems and disordered systems~\cite{mezard2009information,zdeborova2016statistical}.

In analytically tractable models, including linear regression, random features, and kernel methods, the double-descent curve and the associated interpolation peak can be characterized rigorously in terms of spectral properties of data covariances and effective noise levels~\cite{hastie2022surprises,gerace2020generalisation,bordelon2020spectrum,adlam2020neural}. In these settings, the peak arises from the divergence of estimator variance near the point where the empirical covariance matrix becomes singular, providing a precise mathematical counterpart to the notion of criticality. More recent analyses have extended this understanding to the phenomenon of ``benign overfitting''~\cite{bartlett2020benign,tsigler2023benign,frei2022benign}, revealing how high-dimensional systems can absorb (label) noise into high-frequency spectral modes without degrading generalization on the signal component.

For nonlinear deep networks, the situation is more complex. While empirical studies clearly demonstrate double-descent-like behavior across parameters and datasets, a complete theoretical characterization remains elusive. Nevertheless, recent rigorous studies on \textit{benign overfitting}---integrating training dynamics with generalization analysis---have unveiled the tip of the iceberg~\cite{allen2020towards,cao2022benign,jiang2024unveil,chen2023understanding,huang2023understanding}. These works reveal that the transition from \textit{harmful} to \textit{benign} overfitting is critically governed by the interplay between {sample size} and the {signal-to-noise ratio} (SNR). Consequently, the analogy with phase transitions remains conceptually useful: it emphasizes that qualitative changes in generalization behavior can occur as control parameters such as parameters, data size, or regularization are varied, and that these changes are intimately tied to the geometry and dimensionality of the solution set rather than to local properties of individual minima.

\section{Mean-field and dynamical theories of infinitely-wide neural networks}
\label{sec:mean_field}

A fruitful theoretical perspective on modern deep learning emerges in controlled infinite-width limits~\cite{jacot2018neural,mei2018mean,yang2019wide,huang2021towards,huang2021understanding}, which give rise to two complementary descriptions: a kernel regime (``lazy'' training) and a mean-field regime (feature learning). In both cases, analytical simplifications connect neural networks to well-studied objects in probability theory, statistical mechanics, and variational methods. These limits replace random high-dimensional weight configurations by effective macroscopic quantities (kernels or parameter distributions), transforming training into tractable dynamical systems and providing a foundation for understanding the collective behavior of wide neural networks.

\subsection{Neural network Gaussian processes (NNGP)}
\label{sec:nngp}

In the infinite-width limit at initialization, the pre-activations of neurons in each layer become jointly Gaussian by a layer-sequential application of the central limit theorem, provided the weights are independently initialized with zero mean and appropriately scaled variance. Consequently, the network's output converges in distribution to a Gaussian process (GP)~\cite{matthews2018gaussian}. Each forward pass can be viewed as sampling a function from this GP prior, whose covariance kernel depends solely on the network's architecture, activation function, and initialization statistics~\cite{neal2012bayesian,lee2017deep}.

Consider a fully-connected network of width $n_l$ with pre-activations $h^{(l)}(\cdot)$ and activations $x^{(l)}(\cdot)=\phi\!\left(h^{(l)}(\cdot)\right)$:
\begin{equation}
h^{(l)}(x)=W^{(l)}x^{(l-1)}(x)+b^{(l)},\qquad x^{(0)}(x)=x,
\label{eq:fcn_def}
\end{equation}
where $\phi(\cdot)$ is a pointwise nonlinearity. At initialization, we assume i.i.d.\ Gaussian weights with the standard variance scaling
\begin{equation}
W^{(l)}_{ij}\sim \mathcal{N}\!\left(0,\frac{\sigma_w^2}{n_{l-1}}\right),\qquad
b^{(l)}_{i}\sim \mathcal{N}\!\left(0,\sigma_b^2\right),
\label{eq:init_scaling}
\end{equation}
independently across layers and coordinates.

In the infinite-width limit $n_l\to\infty$, the pre-activations become jointly Gaussian and the induced covariance admits a closed recursion. Writing the layer-wise kernel as the covariance of activations, one obtains
\begin{equation}
\Sigma^{(l+1)}(x,x')
= \sigma_b^2 + \sigma_w^2\,
\mathbb{E}_{(u,v)\sim \mathcal{N}(0,\mathrm{Cov}^{(l)}(x,x'))}\!\left[\phi(u)\phi(v)\right],
\label{eq:nngp_recursion}
\end{equation}
where $\mathrm{Cov}^{(l)}(x,x')$ is the $2\times2$ covariance matrix
\[
\mathrm{Cov}^{(l)}(x,x')
=\begin{pmatrix}
\Sigma^{(l)}(x,x) & \Sigma^{(l)}(x,x')\\
\Sigma^{(l)}(x',x) & \Sigma^{(l)}(x',x')
\end{pmatrix},
\]
and the expectation is taken with respect to the random initialization (equivalently, over $W^{(l)},b^{(l)}$). As the depth increases, this recursion defines a hierarchy of kernels that characterize correlations between functions computed by the network.

The NNGP limit provides a clean and analytically tractable description of neural networks as Gaussian process models, enabling exact Bayesian inference via closed-form posterior updates. This framework has been valuable for understanding the inductive biases of architectures~\cite{schoenholz2016deep,yang2017mean,xiao2018dynamical,chen2018dynamical,huang2019mean,huang2021understanding,hayou2019impact,gilboa2019dynamical}.

From a statistical-physics viewpoint, the NNGP regime corresponds to a quenched disorder system: random weights are frozen microscopic configurations that induce a fixed distribution over functions at initialization. Training departs from this static picture in two qualitatively different infinite-width regimes. In the NTK (lazy-training) regime, parameters move only slightly, and learning is well-approximated by kernel regression with an (approximately) fixed tangent kernel. In contrast, the mean-field (feature-learning) regime exhibits nontrivial evolution of the parameter distribution and learned representations, leading to a genuinely nonlinear macroscopic dynamics. Together, these perspectives turn the passive initialization ensemble into an active dynamical system, while emphasizing different mechanisms of learning.

\subsection{Neural tangent kernel (NTK)}
\label{sec:ntk}

Building upon the NNGP framework, the Neural Tangent Kernel (NTK) provides a refined analytical description of the learning dynamics in infinitely wide neural networks~\cite{jacot2018neural}. The NTK formalism considers the evolution of network outputs under gradient descent (GD) or stochastic gradient descent (SGD) when the network is initialized with random weights and trained under the so-called NTK parameterization~\cite{jacot2018neural}. In this limit, as the width of each layer tends to infinity, the empirical fluctuations of the weights vanish, and the network's output dynamics linearize around initialization.

Formally, for a network function $f_{\theta}(x)$ parameterized by weights ${\theta}$, the first-order Taylor expansion around the initialization ${\theta}_0$ yields
\begin{equation}
    f_{\theta}(x) \approx
    f_{{\theta}_0}(x) +
    \nabla_{\theta} f_{{\theta}_0}(x)^{\top} ({\theta} - {\theta}_0).
\end{equation}
Under gradient descent, the parameter updates induce a deterministic evolution of outputs governed by the kernel
\begin{equation}
    \Theta(x, x') =
    \nabla_{\theta} f_{{\theta}_0}(x)^{\top}
    \nabla_{\theta} f_{{\theta}_0}(x'),
\end{equation}
which remains constant throughout training in the infinite-width limit under NTK-parameterization~\cite{jacot2018neural,chizat2019lazy,geiger2020disentangling}. Consequently, the learning dynamics of the network can be expressed as a fixed-kernel gradient flow~\cite{lee2019wide}:
\begin{equation}
    \dot{f}(x, t) = - \sum_{i}
    \Theta(x, x_i)\,
    \nabla_{f(x_i)} \mathcal{L},
\end{equation}
where $\mathcal{L}$ denotes the loss function. This analytical form implies that the evolution of outputs, and hence generalization, can be solved exactly given the initial kernel matrix, providing a powerful theoretical framework for convergence and generalization analysis in overparameterized systems.

In the infinite-width limit, the network at initialization induces an NNGP with covariance $\Sigma^{(l)}(x,x')$, and simultaneously the empirical NTK converges to a deterministic kernel $\Theta^{(l)}(x,x')$~\cite{jacot2018neural}. These two limits are coupled by the recursion
\begin{equation}
\Theta^{(l+1)}(x,x')
= \Theta^{(l)}(x,x')\,\dot{\Sigma}^{(l+1)}(x,x') + \Sigma^{(l+1)}(x,x'),
\label{eq:theta_rec}
\end{equation}
where
\begin{equation}
\dot{\Sigma}^{(l+1)}(x,x')
= \mathbb{E}_{(u,v)\sim \mathcal{N}(0,\mathrm{Cov}^{(l)}(x,x'))}\!\left[\phi'(u)\phi'(v)\right],
\label{eq:sigmadot_def}
\end{equation}
and $\mathrm{Cov}^{(l)}(x,x')$ is the $2\times2$ covariance matrix formed by $\Sigma^{(l)}(x,x), \Sigma^{(l)}(x,x')$, and $\Sigma^{(l)}(x',x')$. Thus, NNGP specifies the function-space prior through $\Sigma$, while NTK augments it with a tangent (derivative) channel $\dot{\Sigma}$ that governs gradient-based learning in the lazy regime.

The NTK formalism has profound theoretical value for the study of deep learning. It provides one of the first rigorous analytical frameworks that links gradient-based optimization with the theory of kernel methods and Gaussian processes. By linearizing the network around its initialization, NTK allows researchers to derive \emph{exact} training dynamics for overparameterized neural networks and to establish provable guarantees of convergence under gradient descent~\cite{arora2019exact,allen2019convergence,du2019gradient,yang2020tensor,zou2020gradient,huang2020neural}. This connection bridges the gap between the nonconvex optimization of neural networks and the convex theory of kernel regression, enabling a mathematically tractable understanding of generalization~\cite{arora2019fine,cao2019generalization,chen2023analyzing}.

Beyond its theoretical elegance, NTK has found substantial utility in practice as a diagnostic and interpretive tool for modern deep learning. Empirically, the spectral properties of the NTK matrix have been used to predict trainability~\cite{huang2021towards}, optimization speed~\cite{arora2019fine,wang2022deep}, and model sensitivity to initialization~\cite{fort2020deep}. The kernel perspective also allows efficient simulation of infinitely wide networks without explicit parameter training, facilitating research on architectural comparisons~\cite{lee2020finite}. Variants such as the empirical NTK have been employed to study transfer learning~\cite{radhakrishnan2023transfer}, and neural architecture search~\cite{chen2021neural}.

From a physical viewpoint, the NTK regime represents a ``lazy training'' limit~\cite{chizat2019lazy}, analogous to a linear response regime in statistical physics. The system responds linearly to perturbations without substantial reorganization of its microscopic states (i.e., neuron activations or features). The NTK acts as a fixed effective coupling matrix mediating the interactions between inputs, analogous to a frozen interaction kernel in mean-field spin systems. This property makes the NTK framework analytically elegant but conceptually limited: since the kernel does not evolve, it cannot capture feature learning---the adaptive reorganization of internal representations that characterizes the expressive power of practical deep networks.

\subsection{Mean-field limits and feature learning}
\label{sec:mean_field_limit}

To transcend the limitations of frozen features in the NTK regime, alternative parameterizations have been proposed, leading to \textit{mean-field} descriptions~\cite{mei2018mean,rotskoff2022trainability,chizat2018global,nitanda2017stochastic}. In this scaling, the empirical distribution of neurons evolves according to nonlinear partial differential equations (PDEs), enabling a rigorous characterization of \textit{representation learning} and self-organization. Unlike the NTK regime, where the kernel remains static, the mean-field limit captures the feedback loop between parameter updates and feature evolution, allowing the internal representations of the network to adapt to data.

Formally, consider a two-layer network of the form
\begin{equation}
    f_{\rho}(x) = \int a\, \sigma(w^{\top}x)\, \rho(a, w)\, \mathrm{d}a\,\mathrm{d}w,
\end{equation}
where $\rho(a, w)$ denotes the joint distribution of the neuron parameters $(a, w)$. Training with gradient flow on the loss $\mathcal{L}(f_{\rho})$ induces a Wasserstein gradient flow in the space of probability measures:
\begin{equation}
    \partial_t \rho_t(a, w) = \nabla_{(a, w)} \cdot\! \big( \rho_t(a, w)\, \nabla_{(a, w)} \Psi(a, w; \rho_t) \big),
\end{equation}
where $\Psi(a, w; \rho_t)$ is an effective potential depending on the current population distribution $\rho_t$. This nonlinear PDE describes the collective evolution of neurons and connects directly to concepts from statistical mechanics, where $\rho_t$ plays the role of a mean-field density evolving under self-consistent forces. Stationary points $\rho^*$ of this flow satisfy a first-order optimality condition for the loss functional; adding entropic regularization, as discussed below, turns this into the minimization of a genuine free-energy functional.

\begin{figure}[tbp]
    \centering
    \includegraphics[width=0.8\linewidth]{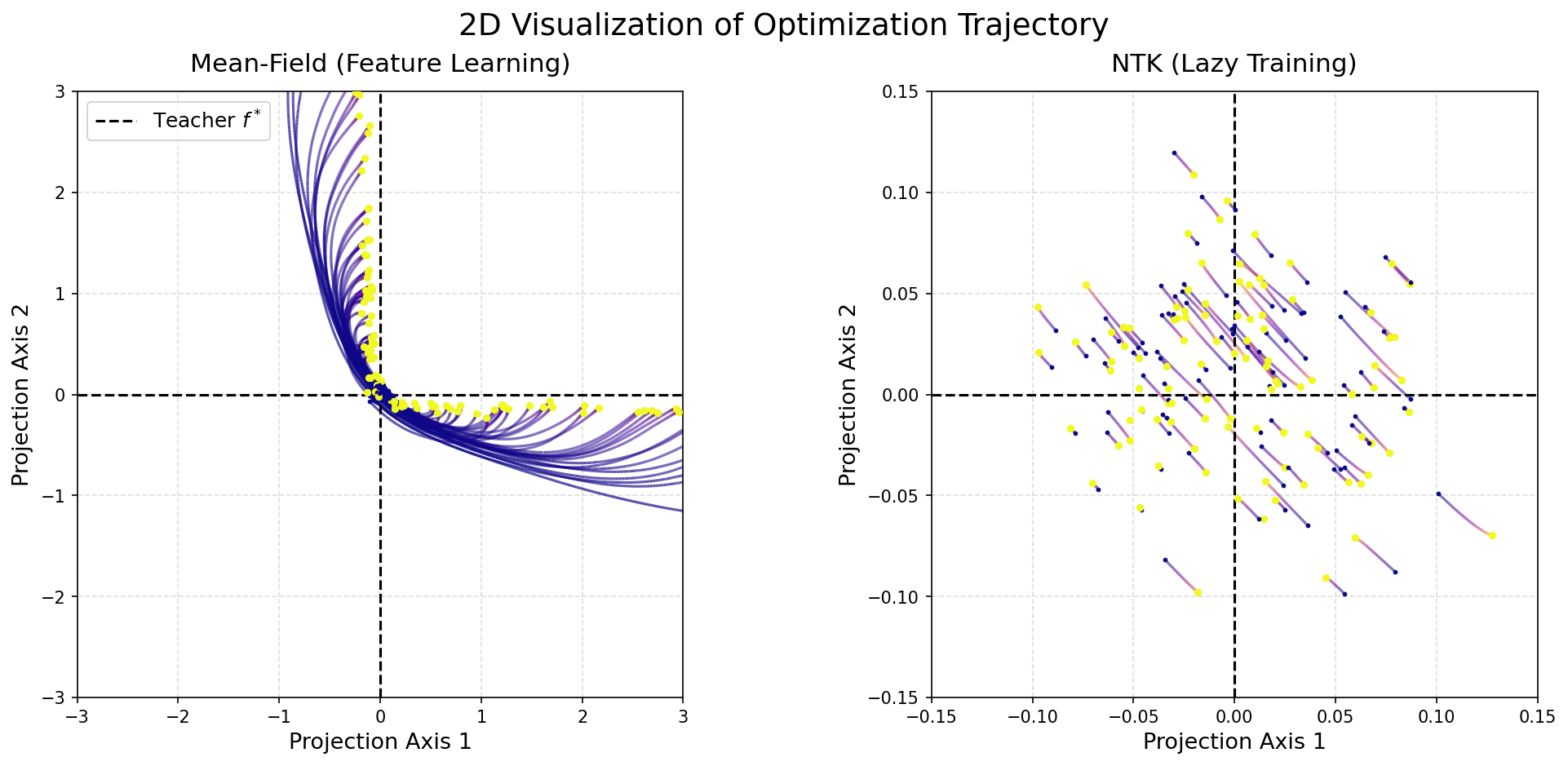}
    \caption{Comparison of optimization trajectories under mean-field and NTK parameterizations in a two-feature teacher model. The first-layer weights are projected onto the teacher $w^\ast_1$ and $w^\ast_2$. Under mean-field training, neurons move far from initialization and align with the teacher directions, indicating feature learning and specialization. Under NTK training, the parameters remain close to initialization, consistent with lazy training and limited representation change. This plot is adapted from~\cite{ba2022high}.}
    \label{fig:mean_ntk}
\end{figure}

The mean-field framework provides a bridge between NTK regimes and the feature-learning regime observed in practice. Under the mean-field scaling---in which the output layer is normalized by $1/N$ rather than $1/\sqrt{N}$ for width $N$, with a correspondingly rescaled learning rate---the network exhibits both tractable analytical properties and rich representation dynamics. In particular, the mean-field equations predict the emergence of structured features and nontrivial attractors in the loss landscape~\cite{chizat2018global,rotskoff2022trainability,mahankali2023beyond}. From a physical perspective, the mean-field limit replaces the fixed kernel of NTK with a self-consistent \textit{interaction field} among neurons, analogous to interacting particle systems or spin glasses with adaptive couplings. Figure~\ref{fig:mean_ntk} provides a simple geometric illustration of the difference between NTK and mean-field regimes. In the NTK regime, the weights stay close to their initialization, so the dynamics are well described by a linearization around the initial network. In contrast, under the mean-field parameterization, neurons move substantially and organize along the teacher directions, reflecting genuine feature learning and the emergence of specialized roles among hidden units.

While the deterministic mean-field equation captures the macroscopic evolution of neuron distributions under gradient flow, a stochastic extension is obtained by adding an explicit isotropic Gaussian noise term to the parameter dynamics. We emphasize that this noise is injected by hand and is structurally different from the data-dependent, anisotropic noise of SGD discussed in Section~\ref{sec:sde}; the resulting Mean-Field Langevin Dynamics (MFLD)~\cite{hu2021mean,chizat2022mean,nitanda2022convex} should therefore be understood as noisy gradient descent rather than as a model of SGD,
\begin{equation}
d\theta_t = - \nabla_{\theta} \Psi(\theta_t; \rho_t)\, dt + \sqrt{2\lambda}\, dW_t,
\end{equation}
where $\theta_t=(a_t,w_t)$ denotes the parameters, $\Psi(\theta_t; \rho_t)$ is the effective potential (drift) depending on the current distribution $\rho_t$, $\lambda$ controls the regularization strength (temperature), and $W_t$ is a standard Wiener process. The corresponding population density $\rho_t(\theta)$ follows the nonlinear Fokker--Planck equation,
\begin{equation}
\partial_t \rho_t(\theta) = \nabla_\theta \!\cdot\!
\big[ \rho_t(\theta)\, \nabla_\theta \Psi(\theta; \rho_t) + \lambda \, \nabla_\theta \rho_t(\theta) \big].
\end{equation}
This stochastic extension describes how noisy gradient descent explores the effective energy landscape under both deterministic drift and isotropic diffusive noise. In the small-noise limit ($\lambda\!\to\!0$), it reduces to pure gradient flow, while finite $\lambda$ enables transitions across metastable states, which may improve exploration. At stationarity, assuming that the effective potential is the first
variation of the energy functional,
$\Psi(\theta;\rho)=\frac{\delta\mathcal{F}_0}{\delta\rho}(\theta)$,
the stationary distribution $\rho^*$ satisfies the self-consistent
relation $\rho^*(\theta)\propto
\exp[-\Psi(\theta;\rho^*)/\lambda]$, and minimizes the regularized free-energy functional
$
\mathcal{F}_{\lambda}[\rho]
=
\mathcal{F}_0[\rho]
+
\lambda\operatorname{Ent}[\rho]$,
where
$\operatorname{Ent}[\rho]
=
\int\rho(\theta)\log\rho(\theta)\,d\theta$. From a statistical-physics viewpoint, the mean-field Langevin equation
defines a self-consistent interacting-particle system whose equilibrium
represents a Gibbs state balancing energy and entropy, linking noisy
optimization with the free-energy formalism of equilibrium statistical
mechanics.

However, unlike classical Langevin dynamics where the drift depends solely on a static external potential, the MFLD drift term depends on the macroscopic distribution $\rho_t$ itself. This nonlinearity historically made global convergence rate analysis exceedingly difficult. Recent progress in the convex analysis of MFLD addresses this difficulty by combining optimal-transport and functional-inequality techniques. The key insight is to introduce the \textit{Proximal Gibbs Distribution} $p_{\rho_t}$~\cite{nitanda2022convex}, defined dynamically as:
\begin{equation}
    p_{\rho_t}(\theta) \propto \exp \left( -\frac{1}{\lambda} \frac{\delta \mathcal{F}_0(\rho_t)}{\delta \rho_t}(\theta) \right).
\end{equation}
The proximal Gibbs distribution $p_{\rho_t}$ acts as a surrogate equilibrium state that tracks the nonlinear drift. By rewriting the nonlinear Fokker--Planck equation in terms of $p_{\rho_t}$, the dynamics mirror those of classical Langevin systems.

Crucially, under suitable convexity and regularity assumptions on
$\mathcal{F}_0$, and assuming that the proximal Gibbs distribution
$p_{\rho}$ satisfies a Log-Sobolev inequality with a constant
$\alpha>0$ uniformly over $\rho$, the entropy-sandwich bound together
with the Log-Sobolev inequality yields a functional
PL-type inequality for
$\mathcal{F}_{\lambda}$~\cite{nitanda2022convex,chizat2022mean}.
Consequently, along the continuous-time MFLD,
\begin{equation}
\mathcal{F}_{\lambda}[\rho_t]
-
\mathcal{F}_{\lambda}[\rho^*]
\leq
\exp(-2\alpha\lambda t)
\left(
\mathcal{F}_{\lambda}[\rho_0]
-
\mathcal{F}_{\lambda}[\rho^*]
\right).
\end{equation}
This establishes exponential convergence of the regularized
free-energy value to its global minimum.

While the infinite-width MFLD admits a clean convergence theory, practical deep learning models are inherently finite. This raises a critical question: how well does the idealized infinite-particle limit approximate a stochastic network with a finite width $N$? In statistical physics, this connection is formalized through the concept of \emph{propagation of chaos} (PoC)~\cite{sznitman2006topics}. It characterizes how particles (neurons) coupled through mean-field interactions become asymptotically independent as the system size $N \to \infty$.

Recent theoretical advances rigorously quantify this finite-particle gap by leveraging Bregman divergences and \textit{Uniform-in-$N$ Log-Sobolev Inequalities}~\cite{suzuki2023convergence,chewi2024uniform,nitanda2024improved}. For a finite-width network trained with noisy gradient descent, the joint distribution of its $N$ neurons at time $t$, denoted $\mu_t^{(N)}$, converges to the optimal mean-field product measure $\mu_*^{\otimes N}$. The gap can be bounded as
\begin{equation}
    \frac{1}{N} \mathrm{KL}(\mu_t^{(N)} \| \mu_*^{\otimes N}) \leq \mathcal{O}\!\left( \exp(-2\alpha\lambda t) \right) + \mathcal{O}\!\left( \frac{1}{N} \right),
\end{equation}
where $\alpha$ is the uniform LSI constant and $\lambda$ is the regularization (noise) scale.

\subsection{Dynamical mean field theory (DMFT) of kernel evolution}
\label{sec:dmft}

To capture representation learning beyond the static NNGP and NTK limits, theorists have recently adapted Dynamical Mean-Field Theory (DMFT)---a mathematical workhorse originally developed for spin glasses and disordered systems---to neural network training~\cite{bordelon2022self,zou2024introduction,mignacco2020dynamical,poc2024dynamical}. Historically employed to characterize collective dynamics, chaos, and memory effects in random recurrent neural networks~\cite{sompolinsky1988chaos}, modern DMFT reduces the coupled, high-dimensional gradient-descent dynamics of millions of weights into an effective, non-Markovian stochastic process for a single ``representative'' neuron.

From a statistical-physics viewpoint, this reduction is naturally formulated using generating-functional methods (such as the Martin--Siggia--Rose path integral) or the equivalent dynamical-cavity method~\cite{zou2024introduction}. In this framework, the complex interactions among neurons are replaced by an effective local field comprising a colored Gaussian noise and a memory retardation term. The macroscopic state of the learning system is entirely captured by a set of dynamical order parameters---specifically, time-dependent \textit{feature covariances}, \textit{response functions}, and \textit{gradient kernels}~\cite{bordelon2022self}. At infinite width, these macroscopic kernels obey self-consistent integro-differential saddle-point equations. This formulation is consistent with the infinite-width limits derived algorithmically via Tensor Programs~\cite{yang2021tensor}, and it suggests numerical schemes that alternate between sampling the effective single-neuron process and updating the kernels, allowing the infinite-width learning dynamics to be solved without instantiating large finite networks.

Conceptually, DMFT complements the dynamical mean-field descriptions discussed in Section~\ref{sec:mean_field_limit}. Standard MF approaches track evolving parameter distributions through nonlinear transport, Fokker--Planck, or integro-differential equations~\cite{sirignano2020mean}. For deep architectures, these distributional descriptions can become increasingly complex and numerically high-dimensional, although the parameter-space dimension need not grow exponentially with depth. DMFT instead adopts an effective-process, or path-space, perspective: under model-specific high-dimensional limits, microscopic degrees of freedom are integrated out to obtain self-consistent equations for representative trajectories and macroscopic order parameters, such as correlation, response, or feature-kernel dynamics~\cite{bordelon2022self,nishiyama2025precise}.

By shifting the fundamental degrees of freedom from weights to features, DMFT tracks multi-time correlation and response functions that encode both deterministic drift and stochastic fluctuations across the entire learning trajectory~\cite{mignacco2020dynamical}. This enables the study of temporal correlations and memory effects that arise during training, producing phenomena analogous to critical slowing down and aging in glassy systems~\cite{zou2024introduction}. Ultimately, by connecting the density dynamics of shallow MF with stochastic, Langevin-type perspectives, DMFT offers a framework that views modern deep learning not merely as descent on a static landscape, but as an interacting, out-of-equilibrium many-body system whose effective geometry and couplings co-evolve with the data.

\section{Learning dynamics as stochastic processes}
\label{sec:sde}

The training of modern deep neural networks is fundamentally a dynamical process evolving through high-dimensional parameter spaces. While Section~\ref{sec:mean_field} focused on the training dynamics of infinite width---where macroscopic observables and parameter distributions evolve according to exact, self-consistent equations---practical training of finite-width networks via Stochastic Gradient Descent (SGD) is driven by intrinsic, finite-batch fluctuations. In this section, we review the statistical-physics framework that treats learning not merely as the evolution of a density, but as a microscopic stochastic trajectory. By drawing parallels between optimization dynamics and the Langevin dynamics of interacting particles in disordered potentials, we explore how state-dependent noise, barrier-crossing phenomena, and non-equilibrium steady states (NESS) actively shape the generalization capabilities of deep networks.

\subsection{Continuum limits and stochastic differential equations (SDEs)}
\label{sec:sde_limits}

A central pillar in the physical analysis of deep learning is the approximation of discrete-time SGD updates as continuous-time Stochastic Differential Equations (SDEs). The standard SGD update rule at step $k$ is given by $\theta_{k+1} = \theta_k - \eta \nabla \mathcal{L}_{\mathcal{B}}(\theta_k)$, where $\eta$ is the learning rate and $\mathcal{L}_{\mathcal{B}}$ is the loss computed on a mini-batch $\mathcal{B}$ of size $S$. Because the mini-batch gradient is an unbiased estimator of the full-batch gradient $\nabla \mathcal{L}(\theta)$, the update can be decomposed into a deterministic drift and a mean-zero stochastic noise term. In the limit of small learning rates, this discrete process can be viewed as the Euler--Maruyama discretization of a Langevin-like SDE~\cite{mandt2017stochastic,li2019stochastic}:
\begin{equation}
    d\theta_t = -\nabla \mathcal{L}(\theta_t)\,dt + \sqrt{\frac{\eta}{S}}\,{C}^{1/2}(\theta_t)\, dW_t,
\end{equation}
where $\mathcal{L}(\theta)$ is the full-batch risk (acting as a physical potential), $W_t$ is a standard Wiener process in the relevant parameter dimension, and ${C}(\theta)$ is the unscaled per-sample gradient covariance matrix. This equation should be understood as a small-learning-rate, diffusion-limit approximation: the drift is the full-batch gradient flow, while the diffusion tensor encodes finite-batch gradient fluctuations whose strength scales as ($\eta/S$).

From a statistical physics perspective, this mapping allows for the application of the Fokker--Planck equation to track the evolution of the probability density $p(\theta, t)$. Unlike classical physical systems in contact with a uniform thermal bath, the effective ``temperature'' in deep learning is explicitly controlled by the hyperparameters: $T_{\mathrm{eff}} \propto \eta/S$~\cite{jastrzkebski2017three,smith2017don,goyal2017accurate,mandt2017stochastic}. This scaling holds in the SDE approximation regime; it neglects momentum corrections (which introduce a factor $1/(1-m)$ for momentum parameter $m$), breaks down at large batch sizes where gradient-descent-like implicit regularization dominates over noise, and may exhibit non-linear dependencies for certain architectures. This correspondence provides a thermodynamic interpretation of training: the drift term $\nabla \mathcal{L}$ drives the system toward low-energy ground states, while the diffusion term ${C}(\theta)$ allows the system to thermally explore the parameter space and escape narrow, sub-optimal local minima.

Furthermore, when the learning rate $\eta$ is finite rather than infinitesimally small, leading-order continuous-time SDE approximations may fail to capture finite-step discretization effects. By employing backward error analysis, discrete gradient-descent dynamics can be approximated by a continuous flow on a modified loss landscape (or effective Hamiltonian). For full-batch gradient descent, the first-order modified loss is $
\widetilde{\mathcal{L}}(\theta)
\approx
\mathcal{L}(\theta)
+
\frac{\eta}{4}
\|\nabla\mathcal{L}(\theta)\|^2$
up to higher-order corrections in $\eta$~\cite{barrett2020implicit}. For mini-batch SGD with random reshuffling, the corresponding modified objective contains additional terms involving the squared norms of the mini-batch gradients, whose effective strength scales with the learning-rate-to-batch-size ratio~\cite{smith2021origin}. Thus, finite learning rates induce an implicit regularization that penalizes large gradient norms and can bias the optimization dynamics toward flatter regions of the loss landscape.

Crucially, if the noise were isotropic and state-independent (${C}(\theta) \approx {I}$), the stationary distribution of this SDE would converge to a classic equilibrium Gibbs distribution $p(\theta) \propto \exp(-\tilde{\mathcal{L}}(\theta)/T_{\mathrm{eff}})$. However, as we discuss in the following subsections, the highly state-dependent and anisotropic nature of SGD noise means that the system often resides in a non-equilibrium steady state (NESS) characterized by persistent probability currents, rather than reaching simple thermal equilibrium.

\subsection{Structure of SGD noise: anisotropy and non-Gaussianity}
\label{sec:sgd_noise}

In classical statistical mechanics, Langevin dynamics typically involve isotropic, additive Gaussian noise, representing a system in contact with a uniform thermal bath. In contrast, the stochasticity in deep learning is inherently state-dependent, highly anisotropic, and structurally coupled to the data. The properties of this noise are crucial for understanding the ``implicit bias'' of SGD toward solutions that generalize well.

A defining feature of SGD noise is its extreme anisotropy. The noise covariance matrix ${C}(\theta)$ is approximately proportional to the empirical Fisher Information Matrix, which in many overparameterized networks aligns closely with the eigenvectors of the Hessian $H(\theta)$~\cite{zhu2018anisotropic,jastrzkebski2017three}. However, this approximation can be poor in practice: the empirical Fisher is a poor proxy for the true Fisher in many realistic settings~\cite{kunstner2019limitations}. Physically, this alignment dictates that the stochastic ``fluctuations'' are strongest precisely in the directions where the ``potential'' (loss) is steepest. This structure induces a self-stabilizing mechanism: the noise actively kicks the system out of sharp minima (where curvature and noise are high) and allows it to settle in flatter regions of the landscape where the noise magnitude diminishes. This behavior constitutes a stark departure from the standard fluctuation-dissipation theorem, as the noise does not merely counteract the gradient descent but actively reshapes the accessible volume of the parameter space~\cite{chaudhari2019entropy}.

A particularly tractable example of state-dependent stochasticity is \textit{label-noise SGD}, in which the labels are independently perturbed during optimization. This mechanism should be distinguished from a dataset whose labels are corrupted once and subsequently held fixed. Near a zero-training-error manifold, the resulting dynamics can exhibit Ornstein--Uhlenbeck-like fluctuations transverse to the manifold together with a slower effective drift along it toward solutions favored by a curvature-dependent implicit regularizer~\cite{blanc2020implicit,damian2021label}. In specific quadratically parameterized models, related label-noise dynamics have been shown to implicitly solve a Lasso problem or favor sparse solutions~\cite{vivien2022label,haochen2021shape}. More recently, the potential benefit of injected label noise has also been analyzed for two-layer networks in binary-classification models with low signal-to-noise ratio~\cite{huang2025does}. Importantly, the interpolation manifold is not generally an absorbing state, because resampled label perturbations can produce nonzero stochastic gradients even when the unperturbed empirical loss vanishes. Thus, the resulting preference for flatter or simpler solutions is model dependent and should not be interpreted as a universal thermodynamic drift toward the flattest available minimum.

Beyond its spatial covariance, the temporal distribution of SGD noise frequently violates the classical Central Limit Theorem. Empirical studies of deep networks suggest that gradient noise often exhibits heavy-tailed distributions, which have been proposed to follow $\alpha$-stable L\'evy processes rather than standard Gaussians~\cite{simsekli2019tail,gurbuzbalaban2021heavy}. However, this characterization remains actively debated: heavy tails may alternatively arise from multiplicative state-dependent noise rather than the additive $\alpha$-stable framework~\cite{hodgkinson2021multiplicative}, and in language models heavy-tailed gradients have been attributed to Zipf's law in token distributions rather than the SGD process itself~\cite{kunstner2024heavy}. Supporting evidence for the L\'evy picture has also been revisited~\cite{battash2024revisiting}. Unlike Gaussian noise, which leads to localized, continuous Brownian diffusion, heavy-tailed noise enables ``L\'evy flights''---discontinuous, long-range jumps in the parameter space. From a dynamical perspective, these jumps can allow the system to leave metastable basins far more readily than under Gaussian Langevin dynamics, with mean escape times governed mainly by the width of a basin rather than by its depth. The tail index $\alpha$ serves as a key dynamical parameter: a smaller $\alpha$ implies more frequent long-range jumps, which has been empirically correlated with the optimizer's ability to traverse the loss landscape and locate wider valleys~\cite{nguyen2019first}.

Together, the noise--curvature alignment, the state-dependent dynamics induced by label noise, and possible non-Gaussian heavy tails indicate that neural network training is generally not well described as simple thermal relaxation. Instead, learning dynamics are better described as a non-equilibrium, driven-dissipative search process, motivating the study of steady-state probability currents which we explore in the next subsection.

\subsection{Non-equilibrium steady states (NESS) and entropy production}
\label{sec:ness}

A fundamental departure from classical equilibrium thermodynamics in deep learning is the violation of detailed balance. In a physical system at thermal equilibrium, the probability flux between any two microscopic states vanishes exactly. However, the state-dependent and non-conservative nature of SGD can violate detailed balance and yield a Non-Equilibrium Steady State (NESS), characterized by persistent probability currents even after the training loss has stabilized.

The evolution of the parameter distribution $p(\theta, t)$ is governed by the Fokker--Planck continuity equation $\partial_t p + \nabla \cdot {J} = 0$, where ${J}(\theta, t)$ represents the probability current. In a strict equilibrium, the stationary state satisfies $\partial_t p = 0$ and ${J} = 0$. In the NESS of deep learning, however, the system reaches stationarity ($\partial_t p = 0$) while maintaining a non-zero flux (${J} \neq 0$). Mathematically, this occurs because the deterministic drift (the loss gradient $\nabla \mathcal{L}$) and the anisotropic diffusion tensor (the noise covariance ${C}(\theta)$) do not perfectly align to form a conservative potential field. Recent theoretical work has shown that these non-zero currents manifest as ``circulatory'' or solenoidal flows in the parameter space~\cite{hu2017diffusion,mingard2021sgd,chaudhari2018stochastic}. Most rigorous treatments of NESS in SGD remain within the SDE approximation; however, exact stationary distributions have recently been derived for diagonal linear networks using the full multiplicative noise structure of SGD, revealing phase transitions, broken ergodicity, and fluctuation inversion that go beyond the standard isotropic Langevin approximation~\cite{ziyin2023law}. Consequently, the optimizer does not merely settle into a static point; rather, it continues to orbit within high-entropy manifolds. From a physical perspective, these persistent currents have been argued to aid exploration: they generate a non-conservative transverse force that can prevent the model from freezing in isolated, ``cold'' regions of the landscape, continuously driving it to explore the extended volume of wide valleys~\cite{chaudhari2018stochastic}.

The system's distance from equilibrium can be rigorously quantified by the Entropy Production Rate (EPR), defined as the rate at which the system dissipates energy into the ``thermal bath'' of mini-batch noise. Within the framework of stochastic thermodynamics, the work performed by SGD can be partitioned into two components: a reduction of the physical energy (minimizing the training loss) and a change in the Shannon entropy of the parameter distribution~\cite{goldt2017stochastic,seifert2012stochastic,ziyin2025neural}. A high EPR during the early phases of training corresponds to the rapid ``melting'' or erasure of the initial random weight configuration. Conversely, a non-zero, steady EPR at late stages signifies a dynamic equilibrium where the model continuously refreshes its internal representations against the stochastic gradients. This thermodynamic view suggests an analogy with Landauer's principle: the distillation of meaningful ``features'' from data and the simultaneous erasure of noise would carry a minimum thermodynamic cost, paid through the dissipation of stochastic gradient fluctuations, although making this analogy quantitative for practical training remains an open problem.

By viewing deep learning through the lens of NESS, we move away from the classical ``mountain-climbing'' metaphor of static optimization. Instead, training can be viewed as a driven-dissipative system where the interplay between energy (loss minimization) and entropy (landscape exploration) is mediated by a continuous, non-equilibrium flux of information and noise.

\subsection{Metastability and first passage dynamics}
\label{sec:metastability}

\begin{figure}[tbp]
    \centering
    \includegraphics[width=\linewidth]{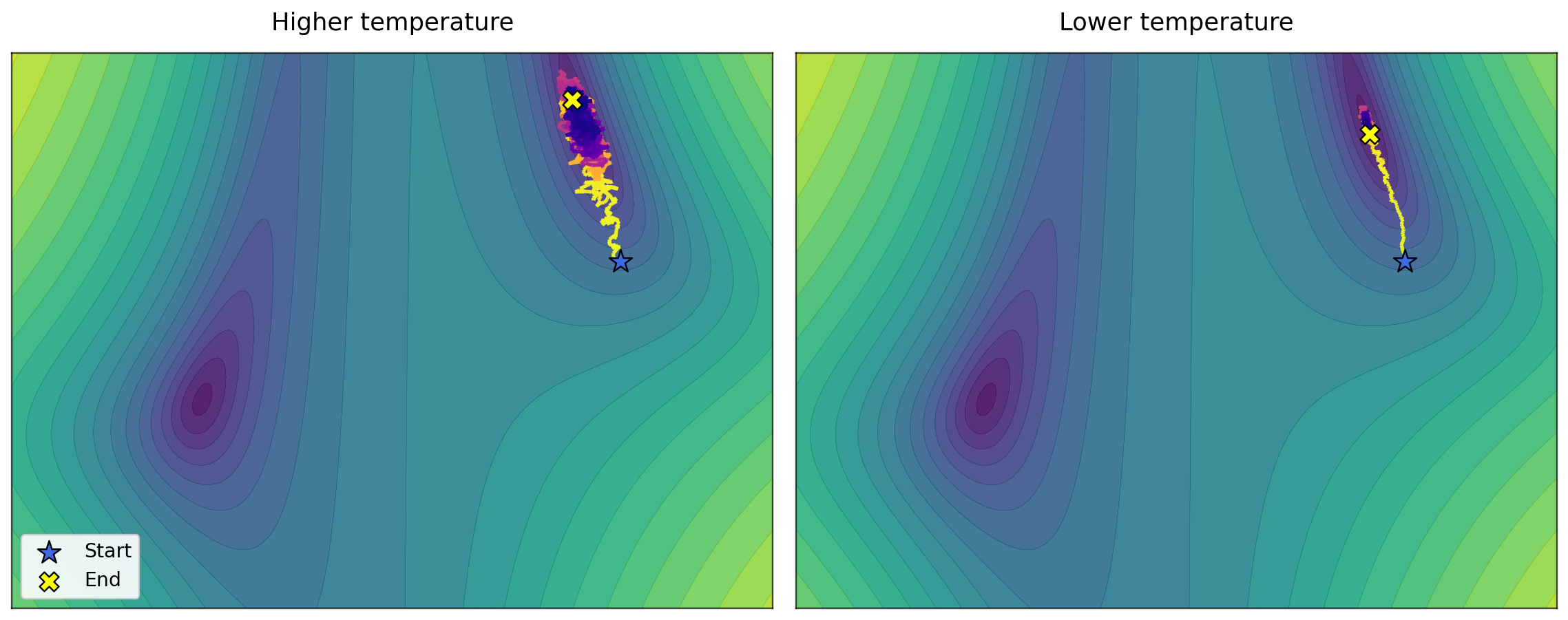}
    \caption{Schematic illustration of temperature-controlled exploration on a non-convex landscape using overdamped Langevin dynamics. Both panels use the same step size, while the effective temperature differs. At higher temperature, the trajectory explores a broader region of the landscape, reflecting enhanced barrier crossing and faster escape from metastable neighborhoods. At lower temperature, the motion remains more localized and stays within the same basin for a longer time. This figure is intended as a qualitative illustration of metastability and first-passage dynamics rather than a faithful simulation of mini-batch SGD.}
    \label{fig:langevin_temperature}
\end{figure}

While the NESS framework describes the steady-state statistics, it is the transient dynamics of barrier crossing that determines the timescale of learning. The loss landscape of deep neural networks is non-convex and, in parts, rugged, with abundant saddle points and possibly metastable regions~\cite{dauphin2014identifying}. In this context, training can be modeled as a problem of metastability: the system must escape from sub-optimal, metastable basins of attraction to reach global or near-global minima.

The timescale $\tau$ required to escape a local minimum is governed by the height of the potential barrier $\Delta \mathcal{L}$. Under the idealized assumption of additive isotropic noise, the mean escape time from a metastable basin would scale in an Arrhenius-like manner $\tau \propto \exp\!\left( \Delta \mathcal{L}/T_{\mathrm{eff}} \right)$. However, this analogy breaks down for SGD, whose noise is multiplicative for the mean-square loss. Mori~et~al.~\cite{mori2022power} showed theoretically that the escape rate is instead governed by the log loss barrier rather than the linear barrier, and depends critically on the number of outlier Hessian eigenvalues---yielding a power-law rather than exponential dependence on the learning rate. This result offers an explanation for why SGD preferentially selects flat minima with low effective dimensionality, providing a rigorous basis for the implicit bias discussed above. A higher learning rate or smaller batch size increases the effective noise level, thereby facilitating barrier crossing. Conversely, as the learning rate decays, the system can be viewed as undergoing a freezing-like crossover, a phenomenon akin to the glass transition in disordered media~\cite{baity2018comparing}. To make this intuition more concrete, Figure~\ref{fig:langevin_temperature} shows a simple two-dimensional schematic based on overdamped Langevin dynamics on a non-convex landscape. Although this is not intended as a faithful simulation of mini-batch SGD, it provides a useful analogue for understanding temperature-controlled exploration. When the effective temperature is high, the trajectory explores a broader region of the landscape and more readily moves along shallow directions and across local barriers. By contrast, at lower temperature, the motion becomes much more localized, and the dynamics remain confined to a metastable basin for a longer time. This qualitative contrast illustrates the basic physical picture behind noise-assisted escape and slow first-passage dynamics.

The stochastic nature of SGD does not explore the parameter space uniformly; it exhibits a strong bias toward ``flat'' minima---regions with low curvature (small Hessian eigenvalues) and high entropic volume. Physically, sharp minima are typically less stable under SGD noise than broader valleys of comparable depth. Due to the anisotropy of the noise, the escape rate from a sharp minimum is significantly higher than that from a wide, flat valley of equal depth. Consequently, the dynamics statistically select solutions that are robust to parameter perturbations~\cite{feng2021inverse}. This bias toward flatter regions is often invoked as one possible explanation for the strong generalization performance of SGD, although the precise notion of flatness and its relation to generalization remain subtle~\cite{keskar2016large,hochreiter1997flat}.

\section{Outlook}
\label{sec:outlook}

Despite rapid progress, a unified statistical-physics theory of deep learning remains incomplete. Existing approaches---including kernel, mean-field, random-matrix, and stochastic-dynamical viewpoints---each explain part of the picture, but their precise relations, regimes of validity, and crossover behavior are still not fully understood~\cite{lee2019wide,mei2018mean,yang2021tensor,pennington2017geometry,martin2021implicit}. Although the distinction between lazy and feature-learning regimes is now well recognized, a quantitative theory of the crossover between them in realistic finite-width settings is still lacking~\cite{lee2019wide,yang2021tensor,mahankali2023beyond,montanari2026phase}.

Another important open question is how to move beyond idealized infinite-width or equilibrium approximations. Many practically relevant phenomena---such as representation drift, neuron specialization, sharpness transitions, and implicit regularization---are finite-width and non-equilibrium effects~\cite{simsekli2019tail,zhu2018anisotropic}. Developing predictive theories for SGD with anisotropic, state-dependent noise remains essential for understanding metastability, flatness selection, and generalization~\cite{li2019stochastic,nguyen2019first,mingard2021sgd,keskar2016large}.

In the mid term, a major goal is to extend these tools to modern architectures, especially transformers, diffusion models, and multimodal systems. While recent works have begun to establish rigorous statistical-physics frameworks for attention mechanisms and transformer dynamics~\cite{jiang2024unveil,tiberi2024dissecting,poc2024dynamical,cui2024phase,rende2024mapping,bordelon2024infinite,rigollet2025mean}, a unified treatment covering the full depth, multimodality, and scale of modern foundation models remains an open challenge. A particularly important direction is to incorporate structured data distributions with latent geometry, correlations, and non-Gaussian features into theory, rather than relying exclusively on isotropic synthetic models~\cite{gerace2020generalisation}.

Looking further ahead, one of the most ambitious questions is whether deep learning admits universality principles analogous to those in statistical physics. Do broad classes of architectures share common phase transitions, scaling laws, or effective macroscopic descriptions~\cite{bahri2024explaining,bordelon2025feature,tamai2025universal}? Can learning dynamics be understood as a form of non-equilibrium many-body self-organization~\cite{mezard2009information,sompolinsky1988chaos,seung1992statistical}? Answering these questions may require deeper cross-talk with condensed matter, non-equilibrium physics, and neuroscience, where related ideas such as criticality, collective computation, and emergent low-dimensional structure have long been studied.

Overall, the most promising outlook is not a single universal theory, but a coherent hierarchy of effective theories that interpolate among the three pillars summarized in Figure~\ref{fig:roadmap}. The central challenge is to connect them along the axes where each idealization breaks down: linking landscape geometry to stochastic dynamics through state-dependent noise, linking infinite-width limits to finite-width corrections through propagation-of-chaos, and linking equilibrium kernel descriptions to non-equilibrium feature learning through dynamical mean-field theory. Building such a hierarchy---from initialization to training dynamics, from feature learning to generalization, and from idealized models to modern large-scale systems---is likely to remain a defining challenge for the field in the coming years.

\begin{acknowledgments}
We thank Zhanpeng Zhou and Liu Ziyin for useful discussion. W.H. is supported by JSPS KAKENHI (24K20848) and JST BOOST (JPMJBY24G6). Y.D. is supported by the National Natural Science Foundation of China (NSFC) under Grants No. 12275263, as well as the Quantum Science and Technology---National Science and Technology Major Project Grant No. 2021ZD0301900. Y.D. is also supported by the Natural Science Foundation of Fujian Province of China (Grant No. 2023J02032).
\end{acknowledgments}

\bibliographystyle{apsrev4-2}
\bibliography{references_cleaned}

\end{document}